\documentclass[11pt]{article}

\usepackage[final]{acl} 
\usepackage{times}
\usepackage{latexsym}
\usepackage[T1]{fontenc}
\usepackage[utf8]{inputenc}
\usepackage{microtype}
\usepackage{inconsolata}
\usepackage{stfloats}
\usepackage{graphicx}

\usepackage{enumitem}
\usepackage{booktabs}
\usepackage[dvipsnames]{xcolor}
\usepackage{outlines,multirow}
\setlist[itemize]{leftmargin=1em,itemsep=0em,parsep=0em,before=\small}
\usepackage[font={itshape,small},rightmargin=0ex]{quoting}
\def\tbullet{$\vcenter{\hbox{\tiny$\triangleright$}}$~}
\usepackage[normalem]{ulem}
\usepackage[framemethod=tikz]{mdframed}

\title{Supporting and Performing Culture from the Inside}

 \author{
  Lea Frermann\textsuperscript{1,2} \and 
  Steven Bird\textsuperscript{3} \\
  \rule{0pt}{3ex} 
  \textsuperscript{1}The University of Tübingen, Tübingen, Germany \\
  \textsuperscript{2}The University of Melbourne, Melbourne, Australia \\
  \textsuperscript{3}Charles Darwin University, Darwin, Australia\\ \\
  \normalsize\texttt{lea.frermann@uni-tuebingen.de\quad steven.bird@cdu.edu.au}
}

\begin{document}
\maketitle
\begin{abstract}
Anthropology and related disciplines which study culture have often found it useful to consider their epistemological and methodological approaches in terms of emic versus etic.
This is the distinction between the insider perspective, how the world is viewed by a member of the culture,
versus the outsider perspective, the scientific cataloguing of cultural knowledge and practice.
We adopt this {framework}
to systematically analyse assumptions in recent research on culture in NLP along the full pipeline of task selection, data collection, system design and evaluation.
We draw attention to the `culture of NLP' which shapes the research focus and approaches of our field,
and suggest pathways towards a more culturally attuned AI.
\end{abstract}

\section{Introduction}
Human culture 
can hardly be disentangled from the culture that a researcher brings to an investigation,
{a phenomenon that has been called \emph{Kulturbrille} (cultural glasses) by Franz Boas 
\citep{Engelke18}.}
The same holds true for the flourishing research on culture and cultural alignment in natural language processing (NLP) \cite{pawar-etal-2025-survey}.
The \emph{culture of NLP} is so deeply embedded that it can be hard to see \citep{Hutchinson24}.
Upstream of the data that flows into NLP systems, culture is the invisible filter that selects, shapes and colours
what is observed, the meaning that is assigned, and the text or speech
that is ultimately {produced} before it is ingested by an LLM.
Downstream of the language technologies we create -- like information retrieval (IR), machine translation (MT), automatic speech recognition (ASR), and most recently, chatbots and large language models (LLMs) --
are all the diverse ways that information and technology are socialised in the world's cultures.

Between the upstream and the downstream {culture that is invisible to NLP,
there is the culture embedded in the representational choices of LLMs
and in the interactional style of chatbots.}
This NLP culture includes evaluations of the capability of {NLP artefacts like} LLMs and chatbots to reproduce culture, including cultural facts but also deeper notions like cultural sensitivity and cultural competence \citep{Nejadgholi26},
or situational appropriateness and values alignment \citep{Minggad26}.
NLP culture is present wherever we limit our conception of culture to information that is documented in textual or spoken records, separated from the
paralinguistic and the extralinguistic,
communicative layers that arise from the embodied nature of communication.
This includes evaluations of the capability of LLMs and chatbots to reproduce cultural facts, but goes further to evaluations of
cultural sensitivity and cultural competence \citep{Nejadgholi26},
or situational appropriateness and values alignment \citep{Minggad26}.

\begin{figure}[t]
  \centering
  \includegraphics[width=0.7\linewidth]{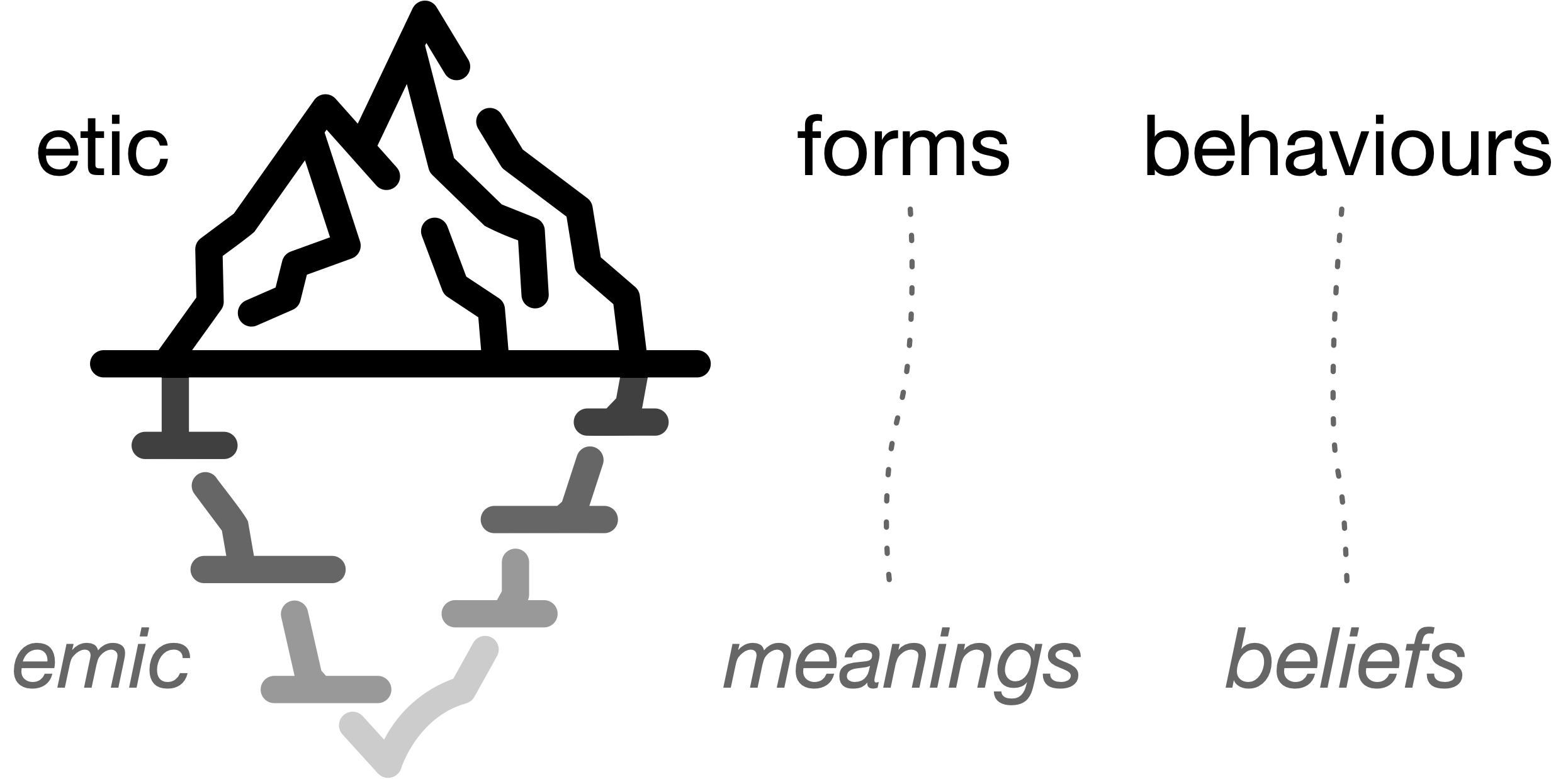}
  \vspace*{-1ex}
  \setlength{\belowcaptionskip}{-3ex}
  \caption{Hall's Cultural Iceberg: forms and behaviours (etic) can be observed by outsiders,
  while corresponding meanings and beliefs (emic) are only accessible to insiders \citep[after][]{Hall76} \label{fig:iceberg}.}
\end{figure}

To systematise epistemological and methodological positions, 
anthropologists have proposed to distinguish emic versus etic approaches to cultural studies~\cite{Pike67etic,Sullivan09}.
An \emph{etic} perspective adopts the outsider's view, classifies surface forms,
and applies generic measures across cultures in a universal, comparative, and scalable view.
The \emph{emic} perspective adopts the insider's view of a member of the culture, with a focus on the underlying cultural system, specific to a single culture.
The concepts of etic and emic originate in structuralist linguistics,
as cultural generalisations of phon\emph{etic} and phon\emph{emic}.
\citet{Pike67etic} observed that human behaviour can be observed, described, and classified
from outside a culture (etic),
while having a significance accessible only to insiders \citep[emic,][]{Headland90}.
A popular analogy for this distinction is the `cultural iceberg' (Fig.~\ref{fig:iceberg}).

The goals of this paper are to
challenge seemingly inevitable methods of research and system development in NLP,
to put forth an alternative perspective,
and to encourage the community to reflect on the culture of NLP and how it impacts our community's goals and contributions.

This paper contrasts emic with etic perspectives across the Four Stages of AI Development, and on this basis
surveys the recent literature on cultural NLP,
identifying a dominance of etic approaches.
It then applies the emic/etic framework and 
(1)~uncovers hidden assumptions in how our community designs approaches, data sets and model evaluations;
(2)~points to gaps in research assuming that the ultimate goal is to develop systems which are valuable for the culture in question; and
(3)~sketches out a path forward towards an emic research agenda.

\section{The Etic/Emic Framework}
\label{sec:culturenlp}

In this section we will see that the practice of generative AI is predominantly etic, in all stages of development.
Task definitions are understood as general purpose solutions to externally-defined problems,
which means that few researchers see any need to engage individual communities about what they need (Sec.~\ref{ssec:task}).
Primary data about the world is taken to be text, yet this contains many layers of culture,
and may effectively be a blind spot for NLP researchers (Sec.~\ref{ssec:data}).
Systems are designed to replicate or replace humans
without due consideration for how this may impact the meaningfulness of lives and livelihoods
on the ground (Sec.~\ref{ssec:system-design}).
Finally, deployments are optimised to serve global measures of efficiency, productivity, and profitability, and performance benchmarks are applied across the board,
regardless of local cultural norms and expectations (Sec.~\ref{ssec:evaluation}).

In addition, we reconsider these AI development stages from an emic perspective.
We incorporate {qualitative} findings from a review of
all papers published in the Findings or Main streams of *ACL and EMNLP conferences (2022--4/2026)
containing the word `culture' or `cultural' in the title (N=107). 
See Appendix~\ref{app:litsurvey} for details of our approach
and a link to a GitHub repository containing the full list of papers and labels.

\subsection{Tasks}
\label{ssec:task}

In the context of a minority culture,
an etic approach usually presumes the presence of the dominant culture,
and looks across cultures, seeking opportunities for universal solutions.
The most obvious need, at this level, is `overcoming language barriers' \citep{Pastor17}.
The default task, then, would be translation.
We can observe this by submitting a queries to a chatbot in progressively
smaller languages and -- beyond a certain threshold --
notice that the chatbot shifts from answering queries in monolingual mode,
to translating queries into the dominant language, usually English.
{In all such translation processes, minority culture content is shaped along
the lines of dominant cultures \citep{Minggad26}.}

If the default etic position is to overcome barriers,
the default emic position {is often} to maintain barriers,
defending cultural diversity in the face of efforts that {seek to} flatten it.
Community leaders enact their self-determination through intergenerational transmission
of cultural practices, strengthening and safeguarding the local lifeworld \citep{Kimmerer13,Harjo19}.

How this manifests on the ground changes from place to place,
but a common linguistic challenge is to transmit ecological knowledge.
For example, parents teach their children about food preparation,
duties to kin, and the associated knowledge of the country and its seasons.
These knowledge practices are increasingly video recorded and shared,
to the point where individuals and communities may have sizeable collections
of untranscribed audio and video in their local vernacular.

This rarely involves universal information access mediated through local language or culture.
Those members are usually already bicultural and bilingual owing to the history of contact.
This means that they are perfectly capable of navigating the {dominant culture}
using the dominant language.
Such is the nature of bilingualism and language shift \cite{Fishman01,Grosjean21}.

When considering NLP tasks from an emic perspective, we do not seek to `understand' or `translate' or `align',
but allow space for diverse local meanings that fall outside the scope of our systems.
For example, language technologies can support navigation of untranscribed audio and video through spoken term retrieval,
enabling human learning of language and culture directly from nontextual
sources \citep{Bird26sigul}.

Insider-driven considerations like the above are barely mentioned in the NLP literature.
The majority of surveyed papers (N=78; 73\%) focussed on the intrinsic goal of assessing or augmenting cultural knowledge in LLMs,
or so-called `trivia-centered' cultural knowledge representations~\cite{Zhou25,oh-etal-2025-culture,Alkhamissi26}.
The remaining papers address downstream tasks ranging from culturally appropriate dialogue over culturally attuned translation to culture-aware IR.
However, the tasks are selected and formalised by the researchers who are embedded in a dominant culture (by upbringing and/or by academic affiliation), typically with the primary goal of generalizability to many cultures in mind (etic).
We did not encounter papers which co-developed a task with a cultural community.

\subsection{Data}
\label{ssec:data}

For most of the world's cultures, the textual record comes from ethnographic observations
that are subject to the cultural conventions of outsiders \citep{Clifford23}.
The record depends on the observable behaviours of insiders,
and what insiders may report about the underlying meanings and beliefs,
reports which may be unreliable or even deceptive \citep{Borofsky05}.
For example, Said argues that everything written about the `Orient'
is a projection of western culture \cite{Said78}.

This problem is not removed by eliminating outsider texts;
there
is the culture of writing itself:

\begin{quoting}\noindent
... writing, wherever it exists, is always only one of several communication channels available to the members of a society. Consequently, the conditions
under which it is selected and the purposes to which it is put must be
described in relation to those of other channels. \citep[p426]{Basso74}
\end{quoting}

There is also the nature of observation, e.g.:
there are no pure facts \citep{Popper35};
differentiating figure from ground is a perceptual act \citep{Schacter09};
measurement involves culturally-informed choices \citep{Walter13};
and `the world as it is' differs from `the world according to data' \citep[p145]{Mitchell20}.
Thus texts are separated from `reality' by three cultural layers:
what is observed; what meaning is attributed; and how this is rendered into text.

{One response to this situation is to}
consult with members of the minority culture in the data collection process rather than relying exclusively on existing texts as ground truth.
A notable fraction of the surveyed literature (19 papers)
drew on community input in concept selection, example creation and/or data validation, where data was typically contributed by members of the target culture which were based at the respective research institution, or via crowd sourcing.
However, we also observe a strong trend of creating `cultural' benchmarks in an at least partially synthetic way (33 papers),
where often concepts and/or context are first generated by LLMs and later filtered by humans in a practice that necessarily distorts the emic, intra-cultural perspective (see discussion above).
The practice of translating English benchmarks was widely criticised and occurred only three times.
31 papers based their concept samples or examples on existing data sets which often spanned many languages (news, social media, Wikipedia, \dots) and imply the concerns laid out above.

\subsection{System design}
\label{ssec:system-design}

The vast majority of language technologies is designed to enhance human information access, through IR, MT, ASR, or LLMs in the form of chatbots.
The construct of chatbots was founded on the model of a human amateur consulting a human expert
\citep{Weizenbaum66,Woods73}.
Studies of the culture of expertise show that it is not to be understood in terms of
information possessed and imparted, but in terms of interaction and ideology.
For each domain there is a culture whereby experts are socialised, evaluated, and authorised,
and there are particular linguistic performances that produce jargon, structure interactions, and animate evidence \citep{Carr10}.
Amateur-expert consultations have a \emph{shape} which is particular to the domain,
e.g.~in medicine with
opening, complaint, examination/test, diagnosis, treatment/advice, and closing \citep{TenHave89}.
Across many domains, better outcomes are linked to interpersonal rapport \citep{Wilson10}.

Set against this, we have the agenda of making chatbots into domain experts \citep{Metzler21},
with the diversion of web queries into chatbot interactions,
in the name of retrieving answers rather than documents.
However, to make this shift is to 
`miss the big picture of why people seek information and
how that process contains value beyond simply retrieving relevant information' \citep[p221]{Shah22}.
Moreover, it risks the replacement of diverse cultures of expert-shaped learning interactions with a
series of prompt-response dyads shaped by novices.

Generative AI is increasingly deployed in agentic systems, designed with the goal of simulating and eventually replacing human experts, from software engineers to academic reviewers. This reflects the culture of efficiency and automation underlying the dominant western view {\cite{Birhane22values}}. An alternative take is to design AI agents to be {\it augmentative}~\cite{bird-2024-must}, to enhance human expertise rather than seeking to replicate it. This approach bodes better with minority cultures' matters of concern (Section~\ref{ssec:system-design}), and may well lead to more socially sustainable AI development more broadly as we discuss in Section~\ref{sec:path}.

Most of our surveyed papers (73\%) focussed on intrinsic evaluation and did not consider a `downstream system' (cf.~Sec.~\ref{ssec:task}).
Those which were more task oriented did not question the design of a potential system, and adopted the predictable designs for IR, classification, or dialogue as established in the {culture of NLP.}

\begin{table*}[t]
{\scriptsize\setlength{\tabcolsep}{0.65\tabcolsep}
\begin{tabular}{r|p{.25\textwidth}p{.21\textwidth}p{.20\textwidth}p{.24\textwidth}}
\multicolumn{1}{r}{} & {\small \textbf{Task}} & {\textbf{\small Data}} & {\small \textbf{System Design}} & {\small \textbf{Deployment / Evaluation}} \\ \toprule
\multirow{3}{*}{\rotatebox[]{90}{\small \textbf{Etic}}}
& generic solns to external problems
& data extraction for LLMs
& simulate or improve on humans
& efficiency and scalability
\\
& \tbullet \emph{generate facts about any culture}
& \tbullet \emph{translated or synthesized common sense benchmarks}
& \tbullet \emph{text chat; adapting to user language; flexibility in tone/register}
& \tbullet \emph{correctness and completeness of responses}
\\ \midrule
\multirow{9}{*}{\rotatebox[]{90}{\small \textbf{Emic}}}
& local agency in shaping their world, identifying problems, exploring solutions
& elicit cultural concepts and logics
& build / augment human capacity in sustaining cultural vitality
& local decision making, maintaining culture and locally meaningful livelihoods
\\
& \tbullet \emph{understanding western concepts which have no translation (e.g. ``risk'')}
& \tbullet \emph{exegesis of western cultural concepts into local languages}
& \tbullet \emph{redesign of text interaction to follow local teacher-led pattern}
& \tbullet \emph{equipped with actionable information, people make better-informed choices}
\\
& \tbullet \emph{transmitting knowledge practices (e.g. food collection/preparation)}
& \tbullet \emph{videos of people demonstrating knowledge practices, sharing lore}
& \tbullet \emph{collection management and navigation, spoken term retrieval}
& \tbullet \emph{measure user-reported utility and safety}
\\ \bottomrule
\end{tabular}
}
\vspace*{-1ex}
\setlength{\belowcaptionskip}{-2ex}
\caption{Etic vs Emic Perspectives on Four Stages of AI Development}\label{fig:etic-emic}
\end{table*}


\subsection{Deployment and evaluation}
\label{ssec:evaluation}

How can we evaluate the success of technology developed from an emic perspective? 
Typical NLP validation paradigms involve benchmarks, baselines,
and scores that can be compared universally, e.g., across methods, languages, and cultures.
In keeping with the understanding of emic as local and culture-specific,
solutions will vary from place to place.
However, emic level evaluation will always center culturally-meaningful ends,
including cultural survival, intergenerational knowledge transmission, communal thriving and individual wellbeing \citep[cf.][]{Nigatu24}. It will be contextualised in concrete application cases and user groups;
and it will involve ongoing evaluation monitoring changes in user needs.

Of the minority of task-oriented papers in our survey,
none deployed a system in a community and attempted an evaluation in an ecologically valid environment.
Validation was primarily based on benchmarks with automated scoring, occasionally augmented with a user study.
However, human studies typically took the form of comparison of culturally attuned system~A with baseline system~B on researcher-defined variables,
and were thus constrained in focus and context, and insensitive to user considerations beyond the predefined variables.

\section{Paths Forward}
\label{sec:path}

How can we navigate into the emic?
One option is to reproduce the four stages of AI development (Sec.~\ref{sec:culturenlp})
within a local culture through monocultural deployments.
Another is to reproduce mainstream methods under local control and idiomatic interaction designs.
A third begins from a relational understanding of intelligence.

\paragraph{Local adaptation.}
Rather than seeking to contribute cultural data to shape mainstream LLMs,
this is a space of deploying language models within local cultures
\citep[cf.][]{Mahelona23,Biju25,Raj25}.
Harms that come from outsider views and from homogenisation towards dominant norms \citep[cf.][]{Peterson25}
are mitigated by restricting training data to
insider sources.
This remains etic because sources are textual,
and they have the limitations described in Section~\ref{ssec:data}.
However, the benefit of
conducting the four stages of AI development
within a single cultural area is that
the cultural assumptions of users are more likely to align with the cultural
tropes that appear in texts provided by cultural insiders.

\paragraph{Local control.}
Here we ask about approaches to the design of AI which
put the local community's aspirations at the centre 
\citep[e.g.][]{BirdYibarbuk24,Cooper24,Markl24}.
These may start with seemingly mundane tasks
such as text entry and literacy \citep{Moshagen23,Pine25}.
Under local control, the interaction design of a chatbot would be reshaped
along the lines of local knowledge transmission.
Perhaps the initiative is held by teachers, not by learners (who have no entitlement to knowledge).
We provide some suggestions in Table~\ref{fig:etic-emic}.

\paragraph{{Relational AI.}}
Navigating into the emic involves diverse understandings of intelligence \citep{Bridle22,Lewis24},
and on approaches not based on the premise of replicating observable human behaviour \citep{Crawford21,Zuboff22}.
This work exceeds one-size-fits-all thinking \cite[cf.][]{RehmWay23,NLLBTeam24},
and essentialisations of culture \citep[cf.][]{Lenard20}.
We expect this to lead to parochial and hybrid solutions \citep[cf.][]{Srinivasan17,Bird26sigul}.
A promising pathway is to develop human resources:
increasing the capacity of western technologists to engage respectfully
with members of diverse cultures,
and increasing the capacity of local communities to conceive language technologies
that serve local agendas.
This involves rethinking the four stages of AI development under diverse epistemologies
\citep[e.g.][]{Kimmerer13,Harjo19}.

\section{Discussion}

In response to constructive feedback,
we address the primary counter-arguments below:

\paragraph{\itshape (a) The etic/emic distinction is presented as near-categorical rather than a spectrum.}

This suggests 
intermediate positions
between outsider and insider, and that a model could be iterated from etic to emic,
or from forms to meanings, or from behaviours to beliefs.
This approach imagines that culture is little more than a set of
idioms that can be learnt \citep{Kabra23,Attia26}
or values that can be aligned with \citep{Cao24,Kabir25}.
See \citet{Zhou25} for a critique of this perspective.

Alternatively, one might wonder about incrementally shifting some of the Four Stages of AI Development (cf.~Tab.~1, columns).
However, in an emic approach, these stages are tightly {coupled},
strongly influenced by the same underlying culture.

Emic and etic were originally proposed as {complementary} standpoints 
on the nature of descriptions of human behaviour~\cite{Pike67etic}.
Much of the work to date is etic rather than emic, and we contend that
{research on culture in NLP}
would benefit from a greater emphasis on emic approaches.

\paragraph{\itshape (b) Systems should embed a `multi-cultural view of knowledge'}

Conceptually, the emic view by definition considers one culture at a time, rejecting the idea of general cross-cultural structures. Practically, large training data sets are always skewed to majority cultures --- quantitatively due to a skew towards majority culture data and/or qualitatively due to short-cut data augmentation methods like translation or generation of additional data. This leads to a dominance of majority culture downstream, compromising the goal of developing technology that is maximally attuned to the needs and goals of the target culture.

\paragraph{\itshape (c) NLP should 
develop/improve the technical capabilities of a system, leaving its integration into specific scenarios to others (e.g., HCI).}

This is an articulation of the ideology that technology arises in a cultural vacuum.
In fact, technological artefacts are social constructs,
a reality that is explored in the discipline of Science and Technology Studies (STS).
For case studies and reflections that relate to linguistic and cultural diversity,
we refer the reader to \citep{Srinivasan17}.

\paragraph{\itshape (d) Incorporate author demographics into the literature survey as part of determining whether an outsider or insider perspective is applied.}

{
Even assuming we could establish author demographics,
there remains the effect that researchers from minority cultures whose work reaches ACL venues
have already assimilated the culture of NLP with its largely etic approaches.
African initiatives such as Masakhane and Lanfrica
focus on compiling public datasets and replicating the results of GenAI from dominant languages.}

\section{Conclusion}

We have applied the etic/emic framework to notions of culture that arise in NLP,
drawing on a systemic review of the cultural NLP literature.
We have described the limits to cultural alignment that are intrinsic to etic approaches,
given the layers of culture that cannot be removed from --
or that fall outside of -- the textual record accessible to LLMs.
We have shown that cultural alignment of LLMs is not `solved' through etic means,
by manipulating the representations of LLMs or the interactional style of chatbots.
The problem is `society-complete' \citep[cf.][p898]{Resnik25},
where society differs from culture to culture.
We have suggested ways forward, towards a more emic paradigm for research in NLP.

\vfil\clearpage
\section{Limitations}
This 
paper was shaped by the authors' own cultural perspectives.
While our literature survey is {\it representative} of research related to culture in NLP, we do not claim that it is {\it exhaustive} and we might have missed a small number of more emic approaches. Still, our main conclusions that etic approaches dominate NLP and that a more emic angle would benefit the utility and value of cultural NLP work still holds.

\section*{Acknowledgements}
We are grateful for the reviewers' constructive feedback.
Frermann is supported by the Australian Research Council Discovery Early Career Research Award (Grant No. DE230100761).
Bird is supported by Australian Research Council Discovery Projects (DP210100228, DP240101952).
{The authors made an equal contribution to the research reported here.}

\bibliography{custom}

\cleardoublepage
\appendix
\setcounter{table}{0}
\renewcommand\thetable{\thesection.\arabic{table}}  

\section{Literature Survey}
\label{app:litsurvey}

Our main paper results are supported by a survey of the recent NLP literature on culture, with the aim to obtain a \textit{representative} picture of the treatment of culture in recent NLP research. Our literature is systematic, but not exhaustive. We discuss a broader follow-up survey in Section~\ref{app:expandedsurvey}.\\
\\
We started by obtaining all papers published between 2022 and April 2026 in the main or findings track of a top-tier NLP conference (*ACL, EMNLP) which mention `culture' or `cultural' in their title. To keep the scope for manual paper inspection manageable, we removed multimodal approaches, and papers whose primary focus is not culture. We reviewed a total of 107 papers. We manually labelled these papers for the following information:

     \paragraph{Primary contribution} (resource, method, analysis, survey)

     \paragraph{Data creation} (predominantly manual, at least partially synthetic, based on a broad background corpus, \dots). Only for empirical papers (i.e., excluding surveys)

     \paragraph{Task} Assessment/improvement of cultural knowledge in LLMs (upstream or intrinsic) vs. task-oriented papers proposing culturally sensitive applications (downstream or extrinsic). Only for empirical papers (i.e., excluding surveys).\\

The full list of reviewed papers and labels is provided on GitHub.\footnote{\url{https://github.com/ColiLea/emic-culture-nlp/}}
Summary statistics underpinning our analyses in Section~\ref{sec:culturenlp} are provided in Tables~\ref{tab:contributions} -- \ref{tab:data}.

\subsection{Expanded Survey}
\label{app:expandedsurvey}
We complemented our survey, retrieving all papers in the ACL Anthology which mentioned the word `culture' or `cultural' in their abstract or title (N=1536). These were infeasible to manually analyse so we instead searched for an expanded list of keywords indicating an emic approach.\footnote{emic, etic, community-centric, community centric, community-centered, community centered, community-focussed, community focussed, community-focussed, community focussed, anthropological, anthropology, insider, outsider, intercultural} 108 papers mentioned at least one of these terms. Even with the very optimistic assumption that all these papers take an emic approach, they would make up for 7\% of our sample. We note that again this is a crude approximation because both the denominator (there will be other papers about culture which did not meet our initial keyword filter) and the numerator (there may be emic papers not mentioning any of our expanded keywords, and many matched papers won't be emic) are noisy. We provide this complement survey as an additional data point, complementary to our more focussed manual analysis.

\begin{table}[h]
    \centering
    \begin{tabular}{lcc}\toprule
    {\bf Category} & {\bf N} & {$\mathbf{\%}$}\\\midrule
Resource&50&0.467\\
Survey&8&0.075\\
Analysis&15&0.140\\
Method&31&0.290\\\bottomrule
    \end{tabular}
    \caption{Distribution over main contributions in papers in our survey.}
    \label{tab:contributions}
\end{table}

\begin{table}[h]
    \centering
    \begin{tabular}{p{4cm}cc}\toprule
    {\bf Category} & {\bf N} & {$\mathbf{\%}$}\\\midrule
Assess / augment cultural knowledge in LLMs&78&0.729\\\bottomrule
    \end{tabular}
    \caption{Fraction of paper with a focus on assessing or augmenting cultural knowledge in LLMs.}
    \label{tab:tasks}
\end{table}

\begin{table}[h]
    \centering
    \begin{tabular}{lcc}\toprule
    {\bf Category} & {\bf N} & {$\mathbf{\%}$}\\\midrule
culture annotators&19&0.178\\
broad data&41&0.383\\
synthetic&33&0.308\\
translation&3&0.028\\
-- / LLM output analysis&4&0.037\\
-- / survey&8&0.075\\
\bottomrule
    \end{tabular}
    \caption{Distribution over data set creation approaches in the papers in our survey. Papers of the last two categories were not included, as they did not contribute data sets or investigated culture in LLM output (rather than human artifacts).}
    \label{tab:data}
\end{table}

\end{document}